\documentclass[conference]{IEEEtran}
\IEEEoverridecommandlockouts
\usepackage{cite}
\usepackage{amsmath,amssymb,amsfonts}
\usepackage{algorithmic}
\usepackage{graphicx}
\usepackage{textcomp}
\usepackage{xcolor}
\usepackage{multirow}
\usepackage{float}
\usepackage{subcaption}
\usepackage{tabularx}
\usepackage[colorlinks=true,citebordercolor=blue]{hyperref}

\usepackage{subfiles}

\usepackage{xspace}

\newcommand{\pinkdot}[1]{\textcolor{magenta}{{\textbullet}\hspace{1mm}#1}}
\newcommand{\bluedot}[1]{\textcolor{blue}{{\textbullet}\hspace{1mm}#1}}
\newcommand{\reddot}[1]{\textcolor{red}{{\textbullet}\hspace{1mm}#1}}
\newcommand{\greendot}[1]{\textcolor{teal}{{\textbullet}\hspace{1mm}#1}}
\newcommand{\Hquad}{\hspace{0.5em}} 
\def\BibTeX{{\rm B\kern-.05em{\sc i\kern-.025em b}\kern-.08em
    T\kern-.1667em\lower.7ex\hbox{E}\kern-.125emX}}
\begin{document}


\title{VL-Fields: Towards Language-Grounded Neural Implicit Spatial Representations\\
}

\author{
\IEEEauthorblockN{Nikolaos Tsagkas}
\IEEEauthorblockA{\textit{School of Informatics} \\
\textit{University of Edinburgh}\\
Edinburgh, United Kingdom \\
n.tsagkas@ed.ac.uk}
\and
\IEEEauthorblockN{Oisin Mac Aodha}
\IEEEauthorblockA{\textit{School of Informatics} \\
\textit{University of Edinburgh}\\
Edinburgh, United Kingdom \\
oisin.macaodha@ed.ac.uk}
\and
\IEEEauthorblockN{Chris Xiaoxuan Lu}
\IEEEauthorblockA{\textit{School of Informatics} \\
\textit{University of Edinburgh}\\
Edinburgh, United Kingdom \\
xiaoxuan.lu@ed.ac.uk}
}

\maketitle

\begin{abstract}
We present Visual-Language Fields (VL-Fields), a neural implicit spatial representation that enables open-vocabulary semantic queries. 
Our model encodes and fuses the geometry of a scene with vision-language trained latent features by distilling information from a language-driven segmentation model. 
VL-Fields is trained without requiring any prior knowledge of the scene object classes, which makes it a promising representation for the field of robotics. 
Our model outperformed the similar CLIP-Fields model in the task of semantic segmentation by almost 10\%. Project page: \texttt{\url{https://tsagkas.github.io/vl-fields/}}.
\end{abstract}


\section{Introduction}
Recently introduced implicit neural fields-based approaches have demonstrated great potential beyond the area of photo-realistic rendering \cite{n3f,frr,clip-fields}. By using the trained parameters of a neural implicit function, coordinates in 3D space can be mapped to different output quantities, including volumetric density \cite{imap}, semantic labels \cite{in-place}, and material rigidity \cite{phys}, etc. 







In robotics, neural fields are naturally an attractive alternative to traditional spatial representations due to their intrinsic properties. First, encoding scene features in the weights of a fully connected multi-layer perceptron (MLP) can be significantly more memory-efficient, compared to using traditional representations like voxels, whose memory requirements grow cubically with the size of the scene. Second, neural fields as representations are disconnected from the scene's resolution, which gives autonomous robots the ability to query the implicit function on-the-fly, only in the areas of interest. Finally, MLPs model continuous functions, which allows for plausible predictions of unobserved regions and gaps, an important feature for incomplete scans from the exploration of unknown environments. Consequently, neural fields are becoming more and more popular for the purpose of creating rich and compact spatial scene representations with the aim of tackling traditional robotics tasks more effectively (e.g., SLAM). 




\begin{figure}
    \centering
    \begin{tabular}{cc}
         \includegraphics[scale=0.33]{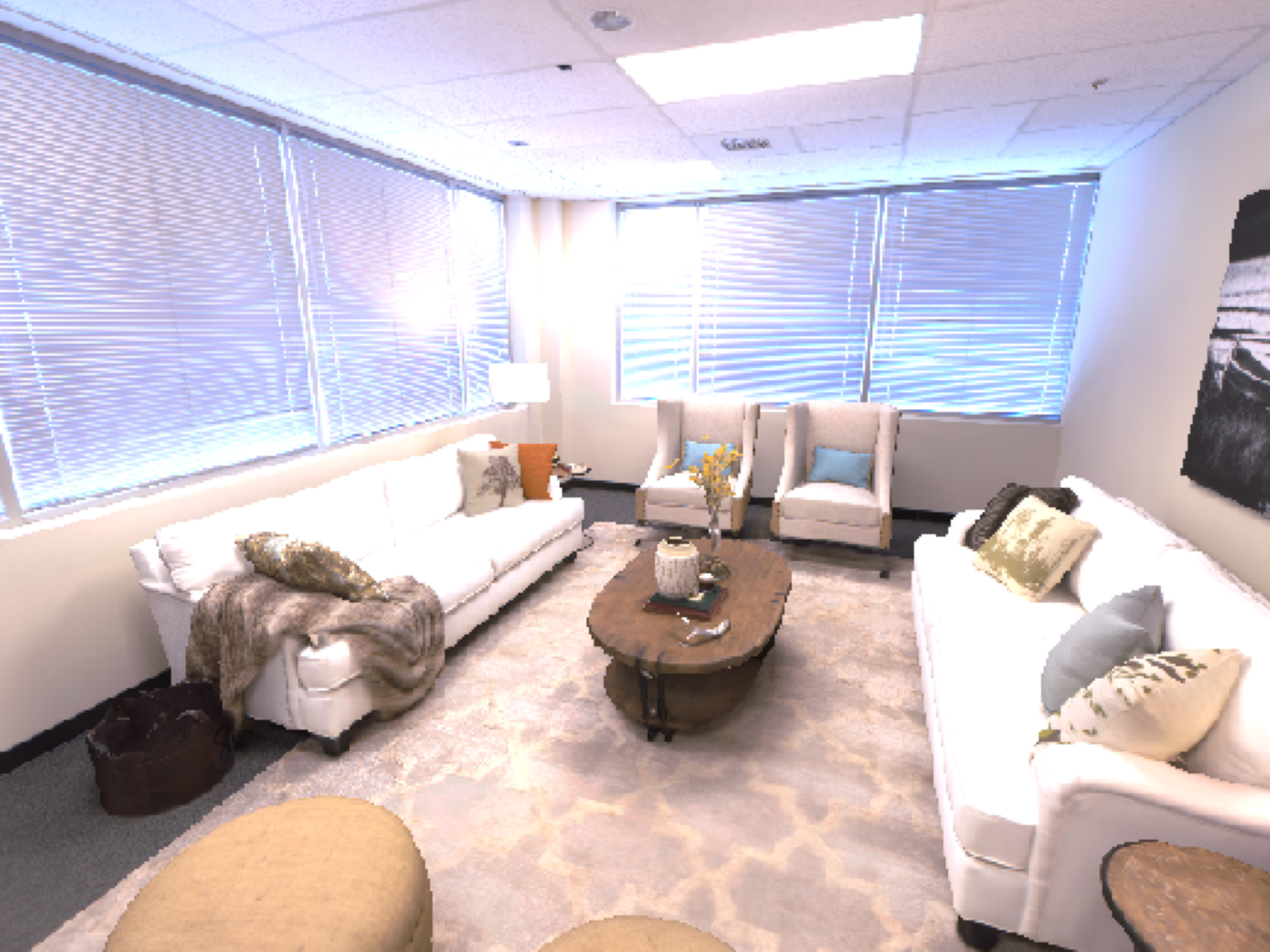}& \includegraphics[scale=0.33]{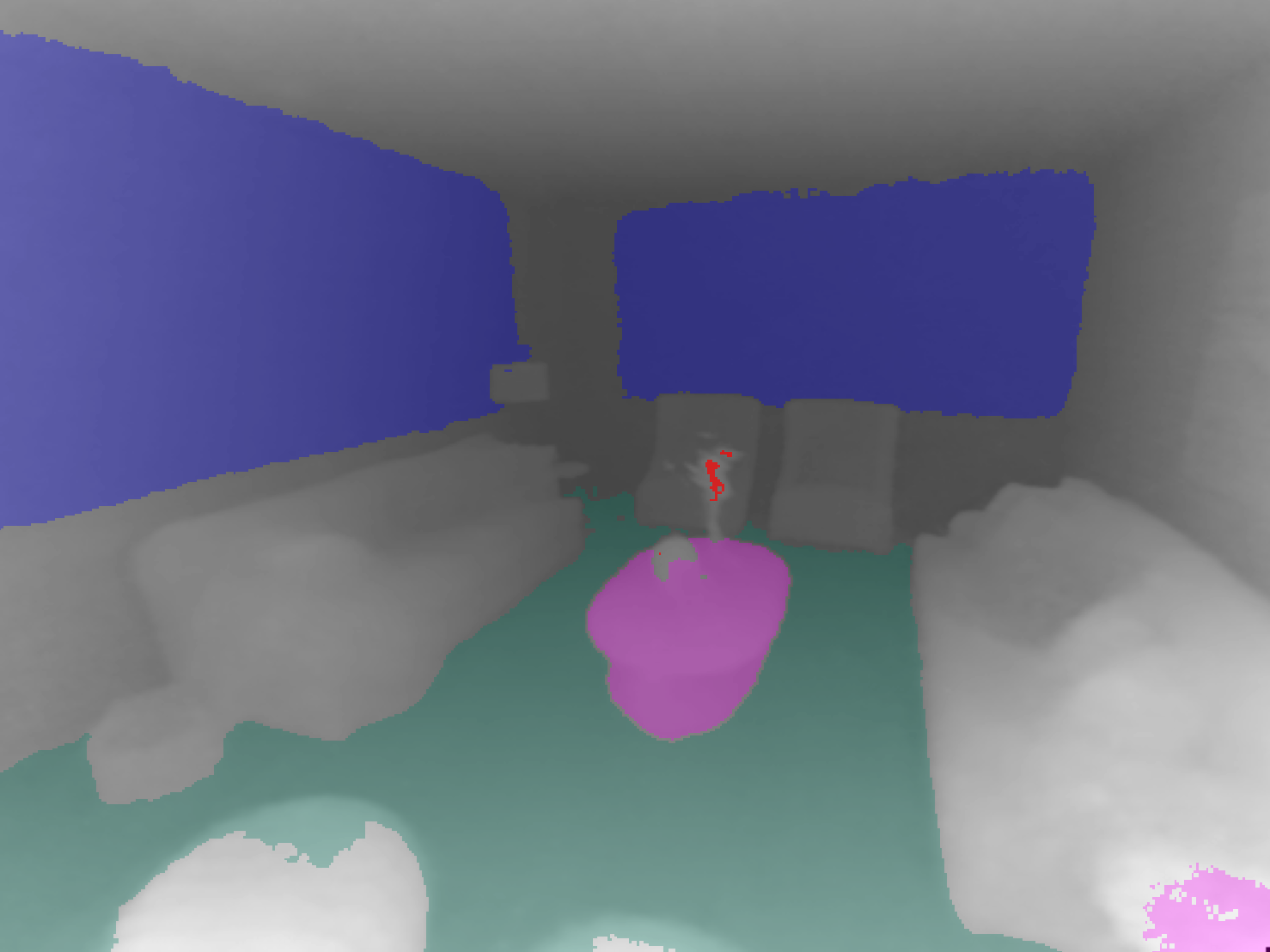}\\
         \includegraphics[scale=0.33]{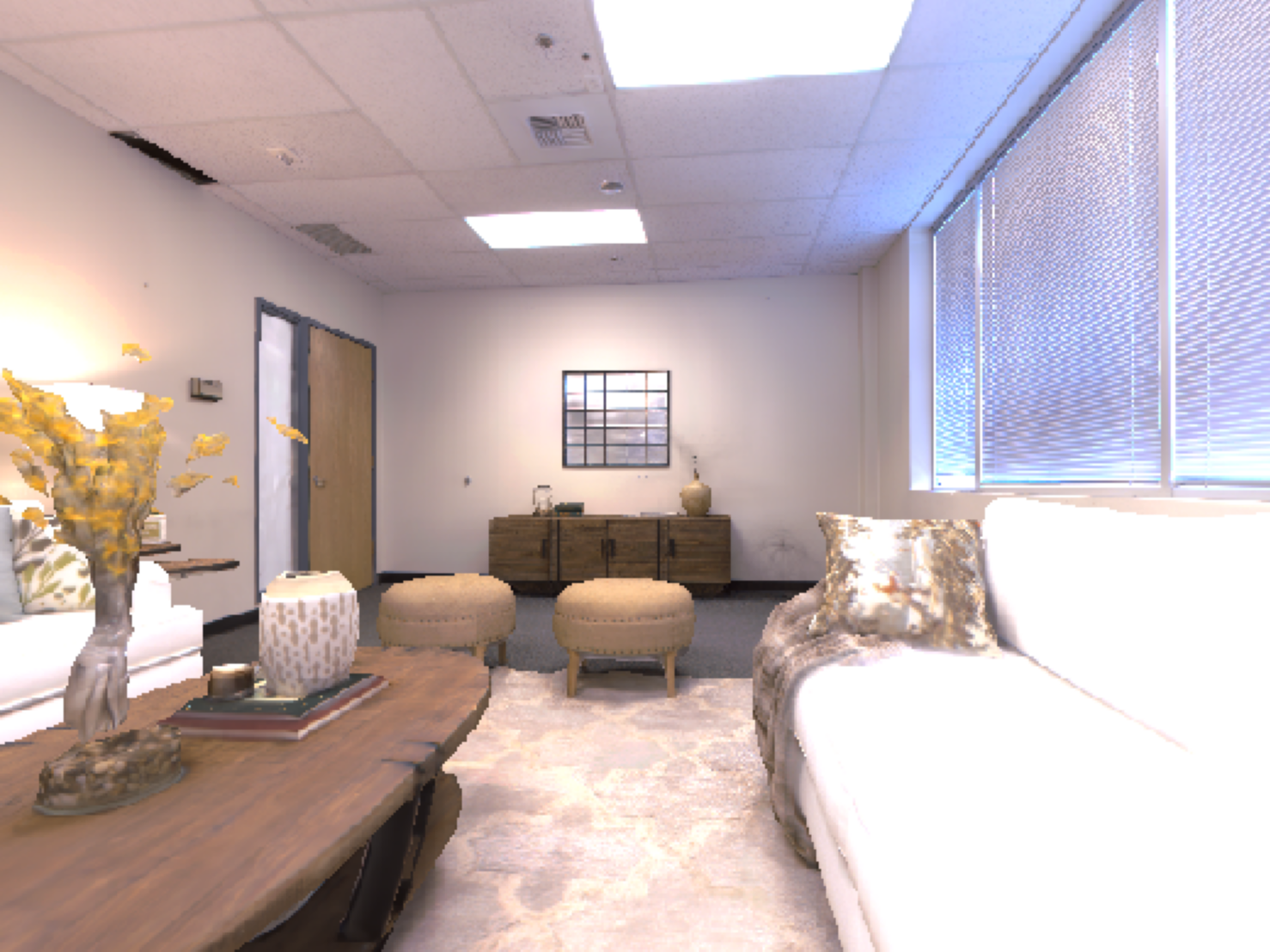} & \includegraphics[scale=0.33]{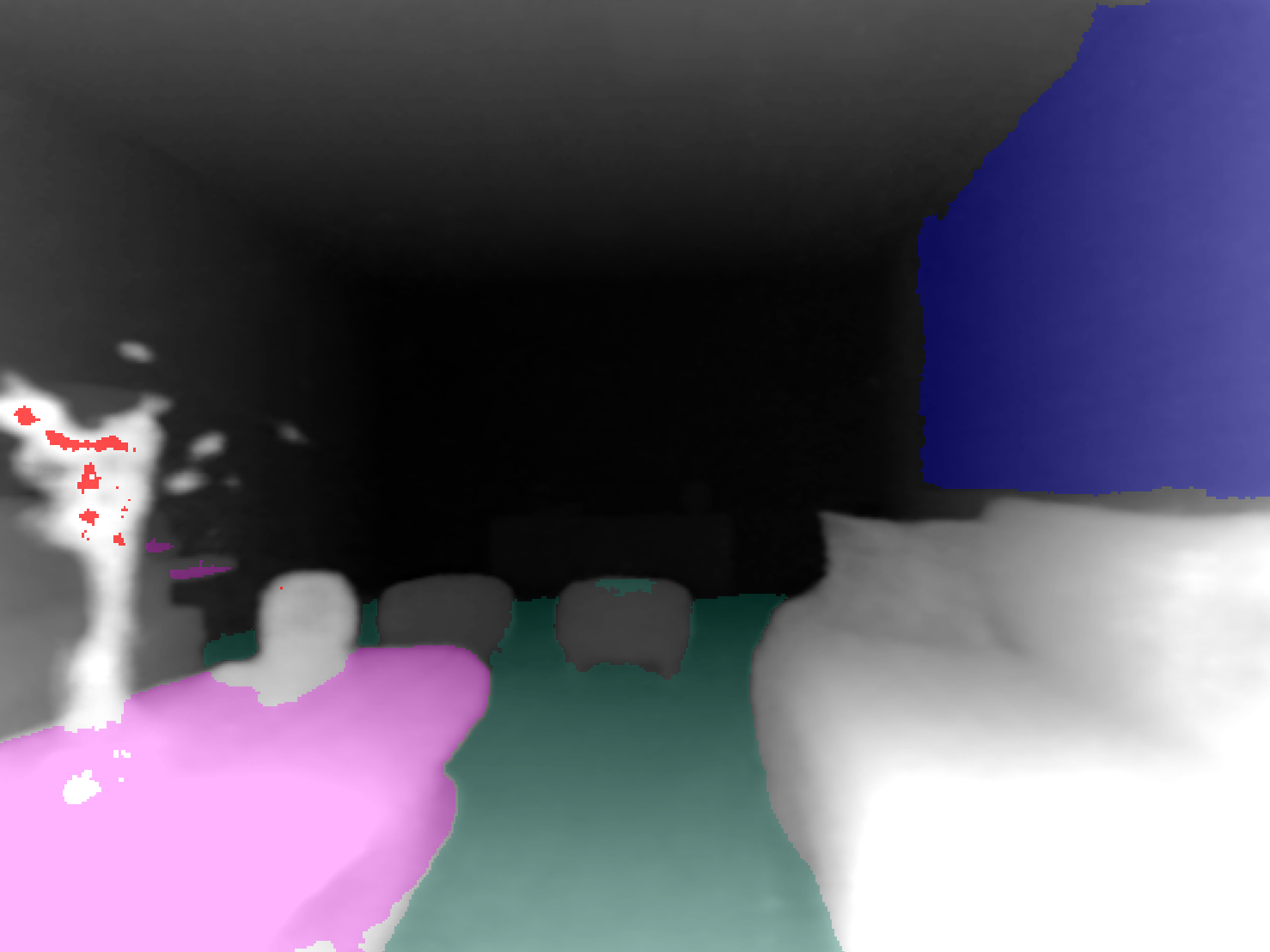}\\
    \end{tabular}
    
    \caption{Our approach grounds open-vocabulary language-based queries in 3D space: \greendot{``vacuum the rug''}, \pinkdot{``clean the table''}, \reddot{``pick up the plant''}, \bluedot{``dust the blinds''}. The colors indicate the areas in the encoded 3D space that correspond to each command.}
    \label{fig:teaser}
\end{figure}

Recently, CLIP-Fields \cite{clip-fields} demonstrated how an implicit function can be trained to map 3D points to high-dimensional embeddings in the CLIP feature space \cite{clip}, where images and texts with similar meanings are represented by vectors that are close to each other. This type of language-grounded neural field can encode the ``semantic memory'' of a mobile robot, thus enabling open-vocabulary queries at run time. For training CLIP-Fields, the off-the-shelf Detic \cite{detic} model was leveraged that provides embeddings in the CLIP space for pixels that correspond to detected objects in an input image. Simultaneously, the text labels of the detected objects were tokenized with sentence-BERT \cite{bert}. Finally, the classified pixels were back-projected to the 3D world and used as input to the network, which was trained with two contrastive losses, one for the label token and one for the visual-language CLIP embedding. 


Nevertheless, this approach has two major limitations. First, CLIP-Fields does \emph{not} encode the geometry of the scene but relies on an external point cloud to perform queries. This choice confines the domain of the implicit function only to points that have been classified by Detic and makes CLIP-Fields restricted to a limited subset of the 3D points of the scene. As a result, we hypothesize that it is difficult for the trained neural field to make reliable predictions for the visual-language features of objects from novel-views, that better capture their geometry. Second, CLIP-Fields assumes that a set of possible scene object classes is available during training, so as to train the model with the contrastive learning paradigm. This assumption potentially limits the ability of the model to execute natural language commands for objects dissimilar to the ones found in the set.  


In our work, we address the aforementioned limitations and propose a new approach for grounding knowledge from visual-language models (VLMs) into neural fields. Our method outperforms CLIP-Fields in the task of semantic segmentation, even though we do \emph{not} require any prior knowledge of the object classes present.

\section{Related Work}
\subsection{Neural Fields}
Neural fields were primarily used for 3D reconstruction tasks (e.g., shape completion) \cite{occ-net, if-gen, deep-sdf}. Due to their ability to handle complex and irregular geometry, the interest in using neural implicit representations quickly spreads to other areas. NeRF \cite{nerf} demonstrated how MLPs can be trained to encode the radiance field of a scene for the purpose of synthesizing photorealistic images from novel views. More recent methods have targeted large-scale scene reconstruction \cite{urf}, faster training \cite{ingp}, dynamic scene encoding \cite{sceneflow-nerf} and more \cite{nf-review}.



\textbf{Semantic Segmentation}: NeRF-based models have demonstrated great accuracy in the task of semantic segmentation. Semantic-NeRF \cite{in-place} and DM-NeRF \cite{dm-nerf} performed scene decomposition, trained with supervision. More recently, NeRF-based panoptic lifting with pre-trained detection models was introduced \cite{panoptic-lifting}, proposing a scheme for dealing with the inconsistent predictions from off-the-shelf models. NeSF \cite{nesf} addressed the lack of generalization, by designing a separate model for performing semantic scene decomposition on numerous scenes that were encoded in different NeRFs. Other techniques used abstract features for the purpose of fusing them with the encoded geometry of the scene, either in the form of activations from off-the-shelf models (i.e., N3F \cite{n3f} and FRR \cite{frr}), or from user interactions (i.e., iLabel \cite{ilabel} and \cite{phys}). Unlike these works, we perform semantic segmentation via neural fields, relying only on open-set vocabulary queries. 




\textbf{Robotics applications}: Several robotics tasks have been explored using the models that encode the geometry of a scene in the weights of a neural implicit function. NeRF-SLAM \cite{nerf-slam}, NICE-SLAM \cite{nice-slam} and iMap \cite{imap} demonstrated how SLAM methods can avail of neural fields and Loc-NeRF \cite{loc-nerf} and iNeRF \cite{inerf} utilized NeRFs for performing pose estimation. Similarly, other works have focused on designing representations that encode the relative position between targets \cite{reaching-distance, neural-desc-fields}. Nevertheless, the use of neural fields in robotics is still in its infancy. 



\subsection{Grounding Language into Spatial Representations}


Web-trained visual-language models have recently managed to encode powerful mappings between images and text, leading to state-of-the-art zero-shot task performance~\cite{clip, lseg, openseg}. This has inspired the grounding of language into spatial representations, for the purpose of enhancing the perception of robots with the ability to perform open-set classification and execution of open-vocabulary queries. In this direction, the augmentation of point clouds has been proposed, via the backprojecting of CLIP embeddings from the pixel domain to the 3D world \cite{vlmaps, open-scene}. This can be achieved with a pre-trained model like LSeg \cite{lseg}, a language-driven segmentation model, which produces per pixel CLIP embeddings. Beyond visual observations, other modalities (e.g., audio) can be used for grounding language  \cite{conceptfusion, avlmaps}.  

However, this strategy can result to memory-expensive semantic maps and the need for visual-language feature fusion schemes, for 3D points that are observed from multiple views. To avoid this, neural fields can be exploited, both for predicting the features, instead of explicitly storing them, and for imposing multi-view consistency, which naturally averages features from many observations. Our model mostly relates to Distilled Feature Fields (DFF)~\cite{ddf} and the more recent CLIP-Fields~\cite{clip-fields}, that train a neural field to predict these embeddings. However, DFF is most applicable for vision and graphics applications, mainly targeting photorealism (i.e., deploys a fine/coarse pair of MLPs, and needs almost a day to converge). Furthermore, its semantic segmentation capabilities are measured only on the points of the ground-truth point cloud, and thus incorrect predictions in empty space are not evaluated. In contrast, our evaluation is conducted per camera ray in the pixel domain, which simultaneously evaluates the quality of the encoded geometry and the grounding of the language features. On the other hand, CLIP-Fields can only make predictions on the classified points of a pre-defined point cloud, thus losing the ability to fuse the detected features with the scene geometry.

\section{Methodology}
We assume access to a collection of posed RGB-D images $I\in\mathbb{R}^{H \times W \times (3+1)}$, depicting different views of an indoor environment. We feed the images to LSeg, which provides a $H \times W \times 512$ feature map in the CLIP embedding space.   

\subsection{Preliminaries}
Following the conventional NeRF rendering approach, we march $N_{R}$ rays $\mathbf{r}$ from the virtual camera's center of projection  through random image pixels $[u,v]$. Along each ray, we select a set of $N$ 3D points via stratified sampling and use the NeRF encoder to predict the corresponding densities $\sigma_i$ and RGB color values $\mathbf{c}_i$. The rendering equation can be approximated via the quadrature rule to estimate the expected color $\hat{C}(\mathbf{r})$ value along each ray:

\begin{equation}
    \hat{C}(\mathbf{r}) = \sum^N_{i=1} T_i(1-\exp{(-\sigma_i\delta_i)})\mathbf{c}_i 
\end{equation}

\noindent where $T_i=\exp{(-\sum_{j=1}^{i-1}\sigma_j\delta_j)}$ and $\delta_i$ is the distance between two adjacent samples. As a result, the colour for each pixel can be approximated as a weighted sum with weights $w_i = T_i(1-\exp{(-\sigma_i\delta_i)})$. We adopt the strategy from N3F, FRR, and DFF, treating VL-Fields as a regular differentiable renderer for all features. As a result, for each ray we can approximate the per pixel estimated color, depth $\hat{D}[u,v]$ and the visual-language feature $\hat{F}[u,v]$ :


\begin{equation}
\label{eq:weighted_sums}
\hat{C}[u,v] = \sum_{i=1}^N w_i\mathbf{c}_i, \Hquad\hat{D}[u,v] = \sum_{i=1}^N w_id_i, \Hquad\hat{F}[u,v] = \sum_{i=1}^N w_i\mathbf{f}_i
\end{equation}

\subsection{Visual-Language Fields}
\label{sub:model}
Similar to CLIP-Fields, our model consists of three components:\\
(1) We utilize multi-resolution hash encoding (MRHE) \cite{ingp} for mapping the input $(x,y,z) \in \mathbb{R}^3$ coordinates to an intermediate 144-dimensional space. 
In contrast to the positional-encoding scheme of the original NeRF work, MRHE allows a neural fields to converge in a small fragment of the time.\\
(2) The outputs are then propagated to a two-layer MLP, where each layer consists of 512 neurons, with ReLU nonlinearities. \\
(3) The MLP's output is passed to two specialized heads. The first predicts the density $\sigma$ and the RGB color value of the corresponding point. 
The other predicts a 512-dimensional vector in the CLIP embedding space. 
    




To train VL-Fields, we minimize the weighted sum of the L2 distances between the predicted and ground truth color (photometric loss $\mathcal{L}_P$), depth (geometric loss $\mathcal{L}_G$), and visual-language embedding (visual-language loss $\mathcal{L}_{VL}$):

\begin{equation}
    \mathcal{L}_{total} = w_P\mathcal{L}_{P}+w_G\mathcal{L}_{G} +w_{VL}\mathcal{L}_{VL}.
\end{equation}

Unlike the original NeRF work, and most of its variants, we do not include the viewing direction $\mathbf{d}$ in the input, as the feature extraction process should be viewpoint-invariant. 


\section{Experiments}
We evaluate VL-Fields on the task of semantic segmentation and compare its performance against both CLIP-Fields and LSeg. Our hypothesis is that the imposed multiview consistency of the neural field will lead to better segmentation accuracy compared to LSeg. We also hypothesize that the encoding of the geometry of the scene will allow our model to fuse the language features to the shapes of the objects, leading to higher quality semantic maps compared to CLIP-Fields. 
\subsection{Experiment Setup}
For our experiments, we use a 5 scenes from the Replica dataset \cite{replica}. We sample 180 posed RGB-D images from each scene and resize the input images to the maximum dimensions LSeg can process (i.e., $H=390, W=520$), which provides us with $H \times W \times 512$ feature maps in the CLIP space.


To train VL-Fields, we set $w_P=1, w_G=0.8, w_{VL}=0.8$ and march 2048 random rays for $10^3$ iterations, sampling 128 points per ray. For training CLIP-Fields, we first generate a dataset by back-projecting pixels to the 3D world with their corresponding CLIP embeddings. Furthermore, following the original setup, we pre-define a set of object labels and classify each point. For this, we use the labels of the objects in each Replica scene. Afterwards, we tokenize the label of each point with sentence-BERT. We train CLIP-Fields for 100 epochs, using 5\% of the point cloud, following the contrastive learning paradigm. Both models require about $50-60$ minutes of training on a mobile RTX 3080Ti GPU.

After training the two models, we sample 45 unseen views for each Replica scene and perform semantic segmentation. For this to happen, we first predict for each sampled pose the per pixel embedding for each model. For VL-Fields, we cast a ray for each pixel and compute the weighted sum as presented in Eq. \ref{eq:weighted_sums}, resulting in a $H\times W\times 512$ visual-language feature map. For CLIP-Fields, we use the depth map to back-project the pixel to the 3D world and query the model at the $(x,y,z)$ coordinates, predicting a visual-language and a label feature map, with $H\times W\times512$ and $H\times W\times716$, respectively. Afterwards, we use the CLIP text-encoder to generate for each Replica label a 512-dimensional embedding. For CLIP-Fields, we also tokenize the labels with sentence-BERT, which results in a 716-dimensional embedding per label. Then, we measure the similarity by taking the dot product between the corresponding vectors and select the class with the highest output. In the case of CLIP-Fields, we follow their setup and compute the weighted sum of the BERT and CLIP dot products, setting the former to be 10 times more important.

\subsection{Quantitative Evaluation}


In Table \ref{tab:miou_segmentation}, the mIoU results are presented for the VL-Fields, CLIP-Fields and LSeg. Our model consistently outperforms CLIP-Fields in all scenes, scoring on average $9.7\%$ higher mIoU. This demonstrates how neural fields benefit from the fusing of high-dimensional features with the encoded scene geometry, even though CLIP-Fields had prior knowledge of the environment object classes. Compared to LSeg, our model scores on average $4.4\%$ higher mIoU. This mainly stems from the imposed multi-view consistency, that implicitly averages predictions from multiple views and adjusts them to the scene geometry.


\subsection{Qualitative Evaluation}
\label{sec:qualitative_eval}
\begin{figure*}[h!]
    \centering
    \includegraphics[width=1.0\textwidth]{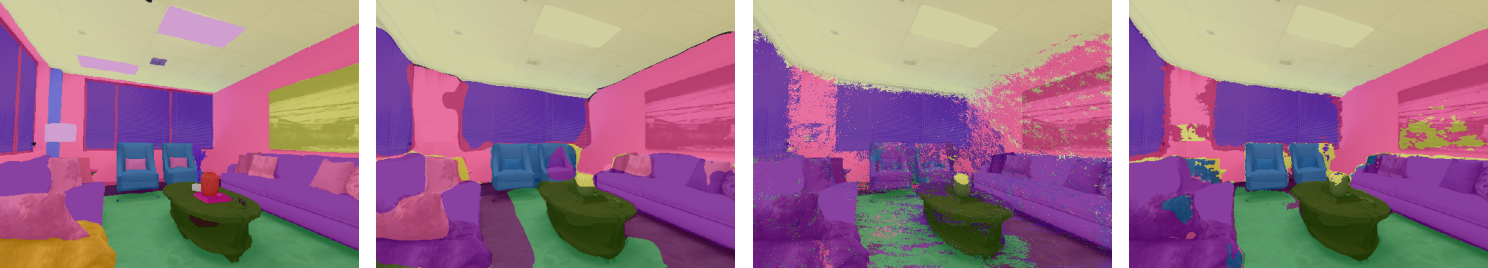}\par
    \vspace{0.2cm}
    \includegraphics[width=1.0\textwidth]{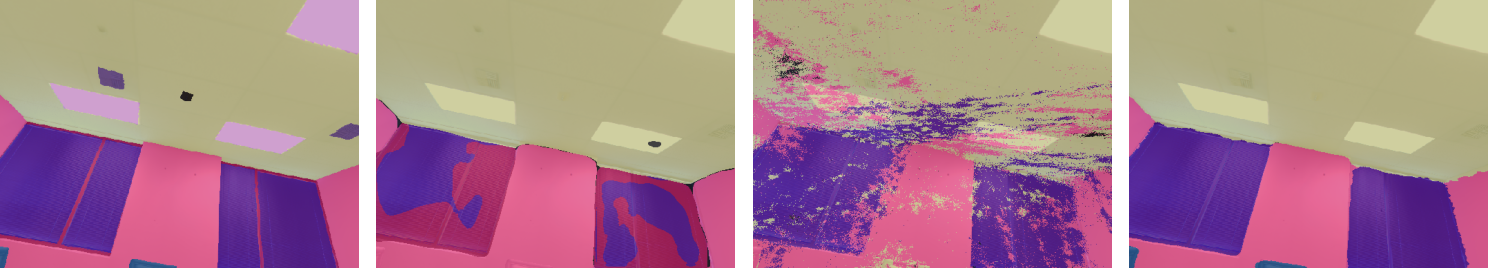}\par
    \vspace{0.2cm}
    \includegraphics[width=1.0\textwidth]{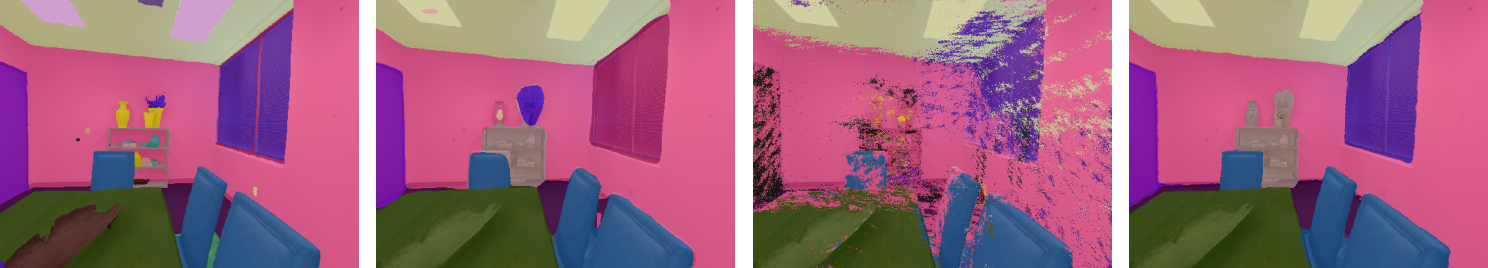}\par
    \vspace{0.2cm}
    \includegraphics[width=1.0\textwidth]{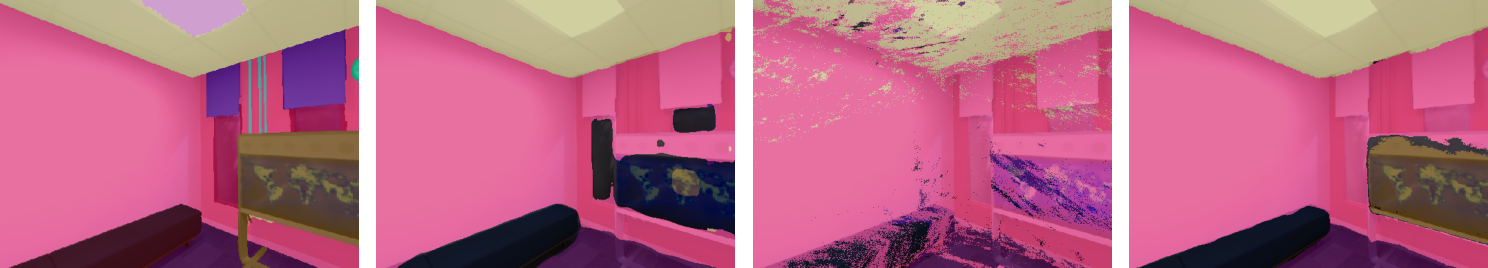}\par
    
    \vspace{0.2cm}
    \begin{minipage}[t]{0.241\textwidth}
        \centering
        \textbf{Ground Truth}
    \end{minipage}
    \hfill
    \begin{minipage}[t]{0.241\textwidth}
        \centering
        \textbf{LSeg}
    \end{minipage}
    \hfill
    \begin{minipage}[t]{0.241\textwidth}
        \centering
        \textbf{CLIP-Fields}
    \end{minipage}
    \hfill
    \begin{minipage}[t]{0.241\textwidth}
        \centering
        \textbf{VL-Fields (ours)}
    \end{minipage}

    \caption{Qualitative comparison between the ground-truth, LSeg, CLIP-Fields, and our VL-Fields semantic maps.}
    \label{fig:qualitative_eval}
\end{figure*}

In Figure \ref{fig:qualitative_eval} we provide four qualitative comparisons between the ground truth, LSeg, CLIP-Fields, and our VL-Fields semantic segmentation predictions. 
Our model is able to filter out many mistakes made by LSeg (e.g., mistaking the floor as a rug and completely missing the painting in the first row). 
Nevertheless, if LSeg consistently makes a specific mistake from multiple views, then both CLIP-Fields and VL-Fields will fail to correct it (e.g., the mis-classification of the bench in the fourth example). 
In general, since VL-Fields fuses features with the learned geometry, we empirically observe overall sharper segmentation (e.g., the windows in the second row). 
Nevertheless, it seems to have a hard time detecting smaller objects, and usually fuses them with the nearest larger object (e.g., the vase and the plant in the third example are both classified as shelfs). 
On the other hand, CLIP-Fields seems to

    


\noindent lack the ability to interpolate effectively and results in overall noisier predictions. We hypothesize that this is stems from the absence of a geometric reasoning. 

\section{Conclusion}
We presented VL-Fields, a novel approach for grounding language into neural fields. Our model consistently outperformed CLIP-Fields and the one-shot LSeg model in semantic segmentation on scenes from the Replica dataset without prior knowledge of the object classes. 
Nevertheless, to a degree, our model still inherits the noisy and inconsistent predictions of LSeg, and seems to perform poorly in recognizing smaller objects. We plan to further investigate instances where our model under-performs, in order to identify the root cause of these issues. We believe VL-Fields is a promising spatial representation for mobile robots, that can act as a compact semantic map that will enable open-vocabulary queries. In future, we aim to evaluate our model for robotics tasks, such as multi-object navigation.

\begin{table}[!hb]
\caption{Semantic segmentation evaluation comparing our VL-Fields (VLF), CLIP-Fields (CF), and LSeg (LS), over all available Replica classes.}

\label{tab:miou_segmentation}
\begin{center}

\begin{tabular}{|c|ccccc|}
\hline
& \verb|room_0| & \verb|room_1| &\verb|room_2| &\verb|office_2| &\verb|office_3| \\
\hline
LS  & 0.559 & 0.583 & 0.736 & 0.740 & 0.752  \\
CF  & 0.515 & 0.593 & 0.720 & 0.699 & 0.681  \\
VLF & \textbf{0.596} & \textbf{0.604} & \textbf{0.810} & \textbf{0.769} & \textbf{0.758} \\
\hline
\end{tabular}
\subcaption*{A. micro mIoU.}
\smallskip

\begin{tabular}{|c|ccccc|}
\hline
& \verb|room_0| & \verb|room_1| &\verb|room_2| &\verb|office_2| &\verb|office_3| \\
\hline
LS  & 0.278 & 0.273 & 0.314 & 0.319 & 0.277  \\
CF  & 0.264 & \textbf{0.292} & 0.334 & 0.256 & 0.275  \\
VLF & \textbf{0.281} & 0.269 & \textbf{0.333} & \textbf{0.359} & \textbf{0.298} \\
\hline
\end{tabular}
\subcaption*{B. macro mIoU.}
\smallskip

\begin{tabular}{|c|ccccc|}
\hline
& \verb|room_0| & \verb|room_1| &\verb|room_2| &\verb|office_2| &\verb|office_3| \\
\hline
LS  & 0.603 & 0.643 & 0.771 & 0.755 & 0.759  \\
CF  & 0.544 & 0.640 & 0.748 & 0.718 & 0.678  \\
VLF & \textbf{0.629} & \textbf{0.657} & \textbf{0.821} & \textbf{0.768} & \textbf{0.761} \\
\hline
\end{tabular}
\subcaption*{C. average mIoU.}

\end{center}
\end{table}

\clearpage
\clearpage
\noindent{\bf Acknowledgments:}
We thank Nur Muhammad (Mahi) Shafiullah for the feedback regarding the training and evaluation of CLIP-Fields. This work was supported by the United Kingdom Research and Innovation (grant EP/S023208/1), EPSRC Centre for Doctoral Training in Robotics and Autonomous Systems (RAS).

\begin{figure}[!h]
    \centering
    \includegraphics[scale=0.2]{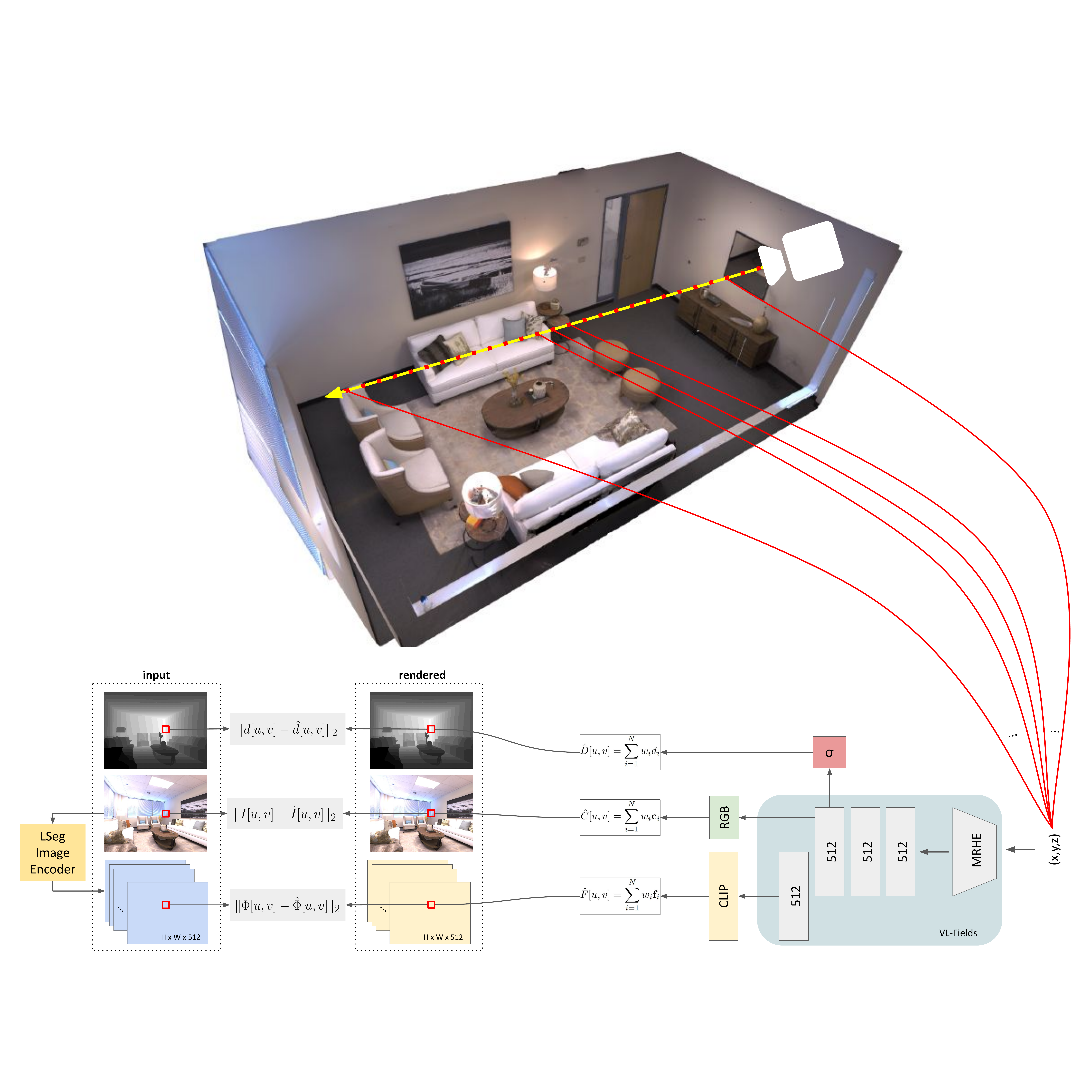}
    \caption{VL-Fields training pipeline.}
    \label{fig:pipeline}
\end{figure}

\clearpage
\clearpage
\onecolumn
\appendix
\noindent In Figure \ref{fig:pipeline}, the architecture of VL-Fields is presented, along with the training pipeline. Rays are marched from different camera poses and the sampled points across each ray are fed to the MLP, which predicts an RGB value, a density $\sigma$ and a $512$-dim embedding in the CLIP feature space. The predictions along each ray are accumulated and the euclidean distance between the ground truth values and the accumulated predictions is computed, and finally backpropagated for updating the weights.

\end{document}